\documentclass{article} 
\usepackage{arxiv}  

\usepackage[utf8]{inputenc} 
\usepackage[T1]{fontenc}    
\usepackage{times}  
\usepackage{helvet} 
\usepackage{courier}  
\usepackage[hyphens]{url}  
\usepackage{graphicx} 
\usepackage{tabularx}
\usepackage{amsmath}
\usepackage{amssymb}
\usepackage{amsthm}
\usepackage{algorithm}
\usepackage{algorithmic}
\usepackage{longtable}
\usepackage{makecell}
\usepackage{stmaryrd}
\usepackage{dblfloatfix}
\usepackage{booktabs}
\usepackage{subcaption}
\usepackage{multirow, multicol}
\usepackage{diagbox}
\usepackage[table,x11names,dvipsnames,table]{xcolor}

\newcommand{\ra}[1]{\renewcommand{\arraystretch}{#1}}

\title{MixPUL: Consistency-based Augmentation for Positive and Unlabeled Learning}
\author{
Tong Wei \\
Nanjing University, China\\
\texttt{weit@lamda.nju.edu.cn}\\
\And
Feng Shi\\
Nanjing University, China\\
\texttt{shif@lamda.nju.edu.cn}\\
\And
Hai Wang\\
4Paradigm Inc., China\\
\texttt{wanghai.ha@gmail.com}\\
\And
Wei-Wei Tu\\
4Paradigm Inc., China\\
\texttt{tuww.cn@gmail.com}\\
\And
Yu-Feng Li\\
Nanjing University, China\\
\texttt{liyf@lamda.nju.edu.cn}\\
}

\def\x{{\boldsymbol x}}

\def\DD{{\mathcal D}}

\def\LL{{\mathcal L}}

\def\NN{{\mathcal N}}

\def\PP{{\mathcal P}}

\def\UU{{\mathcal U}}

\def\WW{{\mathcal W}}
\def\XX{{\mathcal X}}

\newcolumntype{L}[1]{>{\raggedright\arraybackslash}p{#1}}
\newcolumntype{C}[1]{>{\centering\arraybackslash}p{#1}}
\newcolumntype{R}[1]{>{\raggedleft\arraybackslash}p{#1}}
\def\algo{{\textsc{MixPUL}}}

\begin{document}

\maketitle

\begin{abstract}
Learning from positive and unlabeled data (PU learning) is prevalent in practical applications where only a couple of examples are positively labeled. Previous PU learning studies typically rely on existing samples such that the data distribution is not extensively explored.
In this work, we propose a simple yet effective data augmentation method, coined~\algo, based on \emph{consistency regularization} which provides a new perspective of using PU data. In particular, the proposed~\algo~incorporates supervised and unsupervised consistency training to generate augmented data. To facilitate supervised consistency, reliable negative examples are mined from unlabeled data due to the absence of negative samples. Unsupervised consistency is further encouraged between unlabeled datapoints. In addition,~\algo~reduces margin loss between positive and unlabeled pairs, which explicitly optimizes AUC and yields faster convergence.
Finally, we conduct a series of studies to demonstrate the effectiveness of consistency regularization. We examined three kinds of reliable negative mining methods. We show that~\algo~achieves an averaged improvement of classification error from 16.49 to 13.09 on the CIFAR-10 dataset across different positive data amount.
\end{abstract}


\section{Introduction}

\emph{Positive and Unlabeled learning} (PU learning) is emerging in real-world applications since labeling large amounts of data is often prohibitive due to time, financial, and expertise constraints. PU learning typically deals with binary classification and has been applied to novelty or outlier detection~\cite{DBLP:conf/kdd/ElkanN08}, software clone detection~\cite{wei2018ijcai}, and disease gene identification~\cite{yang2014ensemblepu}.

\begin{figure}[tb]
  \centering
  \includegraphics[width=0.5\linewidth]{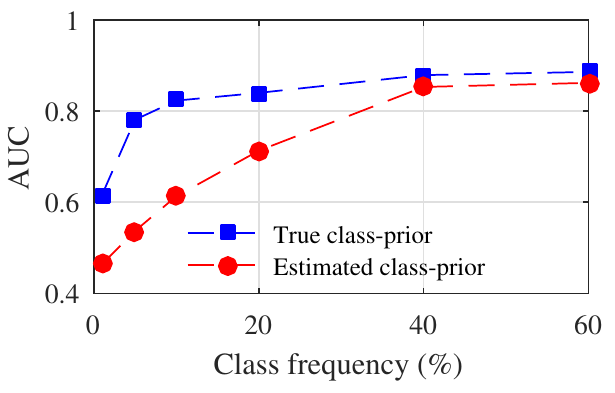}
  \caption{The performance comparison (AUC) using nnPU with true and estimated class-frequency is demonstrated. We vary the class-frequency $c = |\PP| / (|\PP| + \pi |\UU|)$ to simulate PU problems in the wild where $\pi$ is class-prior.}\label{img:estimation-error}
\end{figure}

Given a large number of application scenarios, PU learning has been well studied in recent decades. Previous literature can be divided into two categories based on how unlabeled data is handled. The first line of research is called problem transformation. Through identifying reliable negative examples from unlabeled data, PU learning is transformed into supervised learning~\cite{DBLP:conf/icml/LiuLYL02,li2003ijcai}. Some other work regards unlabeled data directly as negative and considers hidden positive examples among unlabeled data as mislabeled examples. The PU learning problem is transformed into label noise learning~\cite{DBLP:conf/icml/LeeL03,shi2018ijcai}. The second line of research is developing unbiased PU learning risk estimators. This type of research can be seen as cost-sensitive classification~\cite{DBLP:conf/kdd/ElkanN08,DBLP:conf/nips/PlessisNS14,DBLP:conf/icml/PlessisNS15,DBLP:conf/nips/KiryoNPS17}. These unbiased risk estimators typically rely on the knowledge of class-prior which is usually unavailable in real-world problems. Although several approaches have been proposed to estimate the class-prior from PU data~\cite{DBLP:conf/icml/MenonROW15,DBLP:conf/icml/RamaswamyST16,DBLP:journals/ml/PlessisNS17,DBLP:conf/aaai/BekkerD18}, inaccurate estimation usually results in severe performance degeneration as illustrated in Figure~\ref{img:estimation-error}.

It is worth noting that although deep learning achieves excellent performance in semi-supervised learning tasks~\cite{DBLP:conf/nips/SajjadiJT16}, it has not been fully applied to PU learning. Moreover, in much recent work, many effective strategies have been proposed for the training of deep neural networks leveraging unlabeled data, such as \emph{consistency regularization}~\cite{DBLP:conf/nips/TarvainenV17,miyato2018vat} and \emph{mixup}~\cite{DBLP:conf/iclr/ZhangCDL18}. It is demonstrated that these approaches help enhance the performance for semi-supervised learning with a large margin on various problems.

In this paper, we introduce~\algo, a new consistency-based data augmentation algorithm. Unlike previous approaches,~\algo~does not require the knowledge of class-prior. We introduce a unified loss term for PU data that seamlessly improves AUC while encouraging consistency between datapoints. By using mixup, it interpolates pairs of datapoints and their corresponding class labels. The network is then regularized to minimize the distance between its output and the interpolated class labels. It is observed that mixup can move the decision boundary to low-density regions of the data distribution~\cite{DBLP:conf/ijcai/VermaLKBL19} and encourage the model to generalize better to unseen data. 
Due to the absence of negative samples for training, we propose to mine reliable negative examples from unlabeled data to facilitate supervised consistency loss. In extensive ablation studies, we show the effectiveness of consistency regularization and negative example mining techniques.

In summary, our contributions are:
\begin{itemize}
    \item We apply consistency regularization to PU learning to yield a simple yet effective approach which does not need the knowledge of class-prior compared with existing state of the art.
    \item We examine three reliable negative mining methods and show that the randomized technique works best.
    \item We conduct experiments showing that applying consistency regularization can yield substantial improvements over prior state of the art. For example, the proposed method improves classification error from 16.49 to 13.09 on average over the CIFAR-10 dataset.
\end{itemize}

The rest of this paper is arranged as follows. We start by a brief introduction to the problem setting and consistency regularization. Next, we present the proposed algorithm. After that, experimental results are reported followed by the conclusion of this work.

\section{Preliminaries}
\subsection{Problem Setup}
Given $N$ samples $\DD = \{\x_i, s_i\}_{i = 1}^{N}$ where $\x_i \in \mathbb{R}^d$ and $s_i \in \{0, 1\}$. $\x_i$ is regarded as a positive example if $s_i = 1$, otherwise an unlabeled example. We denote the set of positive examples as $\PP$ and unlabeled set as $\UU = \DD \setminus \PP$. The positive set $\PP$ of data is sampled independently from the underlying joint density $p(\x | y = 1)$ and $\UU$ is sampled from a mixture density $p(\x) = \pi p(\x | y = 1) +(1 - \pi) p(\x | y = 0)$ where $\pi$ indicates the \emph{class-prior probability} and $y$ represents the true class label of instance $\x$. The same as conventional supervised learning, PU learning aims to learn a classifier $p(y = 1 | \x)$ which distinguishes positive and negative data.

\subsection{Interpolation Consistency Regularization}
The mixup operator was first introduced by~\cite{DBLP:conf/iclr/ZhangCDL18} for supervised learning and can be defined as:
\begin{equation*}
\operatorname{Mix}_{\lambda}(\x_a, \x_b)=\lambda \cdot \x_a+(1-\lambda) \cdot \x_b,
\end{equation*}
where $\x_a, \x_b$ are two labeled examples and the coefficient $\lambda$ is sampled from the Beta distribution. Later,~\cite{DBLP:journals/corr/abs-1905-02249} adapts mixup to semi-supervised learning and applies interpolation between unlabeled datapoints. Interpolation-based consistency trains a prediction model $f_{\theta}$ to provide consistent predictions at interpolations of unlabeled points:
\begin{equation*}
f_{\theta}\left(\operatorname{Mix}_{\lambda}\left(\x_{j}, \x_{k}\right)\right) \approx \operatorname{Mix}_{\lambda}\left(f_{\theta^{\prime}}\left(\x_{j}\right), f_{\theta^{\prime}}\left(\x_{k}\right)\right),
\end{equation*}
where $\x_j, \x_k$ are a pair of unlabeled examples and $\theta^{\prime}$ is a moving average of the network parameter $\theta$. The interpolation-based consistency can be seen as encouraging the model to have strictly linear behavior ``between'' examples, by requiring that the model's output for a convex combination of two inputs is close to the convex combination of the output for each individual input. The mixup regularizer and consistency loss have not been previously investigated in PU learning and it is interesting to investigate its efficiency.


\section{Method}

\subsection{Consistency Regularization for PU Learning}
During training phase of neural networks $f_\theta$, given a batch $\XX_{\PP} \subset \PP$ of positively labeled examples with corresponding targets and an equally-sized batch $\XX_{\UU}  \subset \UU$ of unlabeled examples,~\algo~produces a processed batch of augmented unlabeled examples $\XX_{\UU}^{\prime}$ with ``soft'' labels using mixup.  $\XX_{\UU}^{\prime}$ is then used in computing the unsupervised consistency loss term. For each pair of two unlabeled samples and their ``soft'' labels $(\x_j, \hat{y}_j), (\x_k, \hat{y}_k)$ where $\hat{y}_j = f_{\theta^{\prime}}\left(\x_{j}\right), \hat{y}_k = f_{\theta^{\prime}}\left(\x_{k}\right)$, an augmented unlabeled datapoint $(\x^\prime, \hat{y}^\prime)$ is obtained as follows using mixup operator:

\begin{equation}
    \x^{\prime} = \lambda \x_j + (1 - \lambda) \x_k
\end{equation}
and
\begin{equation}\label{equ:mixlabel}
    \hat{y}^{\prime} = \lambda \hat{y}_j + (1 - \lambda) \hat{y}_k,
\end{equation}
where
\begin{equation}
\lambda \sim \operatorname{Beta}(\alpha, \alpha)
\end{equation}
and $\alpha$ is a hyperparameter of the Beta distribution. In our implementations, we first collect unlabeled examples with their guessed labels into:
\begin{equation}
    \XX_{\UU} = \{(\x_1, \hat{y}_1), \dots, (\x_B, \hat{y}_B)\}.
\end{equation}
and after performing mixup we get:
\begin{equation}
    \XX_{\UU}^{\prime} = \left\{\bigl(\mathrm{Mix}\left(\x_i, \x_{r_i}\right), \mathrm{Mix}\left(\hat{y}_i, \hat{y}_{r_i}\right)\bigr)\right\}_{i = 1}^{B},
\end{equation}
where $r$ is a random permutation of $[B]$. 
Notably, instead of predicting soft labels $\hat{y}_j$ and $\hat{y}_k$ in Equation (\ref{equ:mixlabel}) using network $f_{\theta}$, we maintain a moving average $\theta^{\prime}$ of parameter $\theta$ following~\cite{DBLP:conf/nips/TarvainenV17,DBLP:conf/ijcai/VermaLKBL19} and set $\hat{y}_j = f_{\theta^\prime}(\x_j)$. Then, we perform $\WW = \text{Shuffle}(\XX_{\UU})$ which will serve as a data source for mixup. For each the $i$-th example and label pair $(\x_i, \hat{y}_i) \in \XX_{\UU}$ and $(\x_{r_i}, \hat{y}_{r_i}) \in \WW$, we apply the mixup operator  and add the result to the collection $\XX_{\UU}^{\prime}$. Note that the interpolation is only applied between unlabeled datapoints so far and on each mini-batch we sample a random $\lambda$ from $\mathrm{Beta}(\alpha, \alpha)$ for mixup. To summarize, by using mixup, $\XX_{\UU}$ is transformed into $\XX_{\UU}^{\prime}$, a collection of multiple augmentations of each unlabeled example with corresponding ``soft'' label.

\subsection{Reliable Negative Mining}
It is noteworthy that applying mixup requires reasonably good ``soft'' labels, which is realized by training networks on labeled data in semi-supervised learning.
In PU learning, it is unrealistic to train the networks by feeding only positive data. We alleviate this problem by identifying a subset of reliable negative (RN) examples from the unlabeled set. In this work, three different types of methods are considered.
\begin{itemize}
    \item \emph{Rand:} We construct a set of ``pseudo'' negative examples by randomly downsampling the unlabeled set.
    \item \emph{Dist:} Unlabeled instances with the farthest averaged distance from positive data are selected as negative.
    \item \emph{NTC:} An Non-Traditional Classifier (NTC) is trained to discriminate $\PP$ and $\UU$. Instances with the smallest prediction scores are selected as negative.
\end{itemize}
The positive set $\PP$ and the selected reliable negative samples $\NN \subset \UU$ are used to compute supervised consistency loss. Even when combining a positive sample and a false negative sample the loss computed can still be useful as the positive sample contains the true label of the other one. It is noteworthy that~\algo~does not reduce cross-entropy loss in case of overfitting. By leveraging consistency training, it can better explore the data space even when labeled data is scarce. We compare three RN mining methods in the experiments. 

\subsection{Objective Function}
We describe each part of~\algo's objective function in the following.
Given batch $\XX_{\PP}$, $\XX_{\NN}$, and $\XX_{\UU}$, we construct $\XX_{\PP\NN}^{\prime}$ by applying mixup operator on $\XX_{\PP}$ and $\XX_{\NN}$. Similarly, $\XX_{\UU}^{\prime}$ is formed by mixing unlabeled data $\XX_{\UU}$. We then combine the consistency loss and the margin loss. More formally, the combined loss $\mathcal{L}$ for our proposed~\algo~is computed as Equation~(\ref{equ:obj}):

\begin{equation}\label{equ:obj}
\begin{aligned}\mathcal{L} &=\mathcal{L}_{\mathcal{PN}}  + \beta \mathcal{L}_{\mathcal{U}} +  \gamma \mathcal{L}_{\mathcal{PU}}. \end{aligned}
\end{equation}

The first two terms respectively represent the supervised and unsupervised interpolation-based consistency loss and can be written as:
\begin{equation}\label{equ:sup-consistencyloss}
\begin{aligned} \mathcal{L}_{\mathcal{PN}} &=\frac{1}{\left|\mathcal{X}_\mathcal{PN}^{\prime}\right|} \sum_{\x, \hat{y} \in \mathcal{X}_\mathcal{PN}^{\prime}}\left\|\hat{y} - f_{\theta}(\x)\right\|_{2}^{2}
\end{aligned}
\end{equation}

\begin{equation}\label{equ:consistencyloss}
\begin{aligned} \mathcal{L}_{\mathcal{U}} &=\frac{1}{\left|\mathcal{X}_\mathcal{U}^{\prime}\right|} \sum_{\x, \hat{y} \in \mathcal{X}_\mathcal{U}^{\prime}}\left\|\hat{y} - f_{\theta}(\x)\right\|_{2}^{2}
\end{aligned}
\end{equation}
By imposing consistency loss, it regularizes the network to have strictly linear behavior. Using ``soft'' labels can also alleviate the problem of absence of negative data.

Since the consistency terms function as regularizers, we introduces a risk function, i.e., margin loss, between pairs of positive and unlabeled samples, which takes the following form:
\begin{equation}\label{equ:pairwiseloss}
\begin{aligned} \mathcal{L}_{\mathcal{PU}} &=\frac{1}{\left|\mathcal{X}_{\mathcal{P}}\right| \cdot \left|\mathcal{X}_\mathcal{U}\right|} \sum_{\substack{\x_{p} \in \mathcal{X}_{\mathcal{P}}, \x_{u} \in \XX_{\UU}}} \left|f_{\theta}(\x_u) - f_{\theta}(\x_p) + \eta \right|_{+},
\end{aligned}
\end{equation}
where $|z|_{+}$ returns $z$ if $z > 0$, otherwise $0$. $\eta$ is the margin parameter. By imposing $\mathcal{L}_{\mathcal{PU}}$,~\algo~is desired to produce higher prediction score for positive samples than unlabeled samples. It is shown that $\mathcal{L}_{\mathcal{PU}}$ can be viewed as an estimation of its supervised counterpart.

Finally, we use hyperparameters $\beta$ and $\gamma$ to trade-off these three terms. When optimizing Problem~(\ref{equ:obj}), we compute the gradient $\nabla_{\theta} \LL$ and update $\theta$ using standard SGD or Adam. Then we update the exponential moving average $\theta^{\prime}$ of network parameter $\theta$ following~\cite{DBLP:conf/nips/TarvainenV17}. 

\subsection{Theoretical Interpretation of Equation~(\ref{equ:obj})}
We further explain our objective function from the perspective of empirical risk minimization.
If we regard the consistency loss $\mathcal{L}_{\mathcal{PN}}$ and $\mathcal{L}_{\mathcal{U}}$ in Equation~(\ref{equ:obj}) as regularizations, the last term $\mathcal{L}_{\mathcal{PU}}$ can be interpreted as a risk function. The consistency loss can move the decision boundary to low-density regions of the data distribution~\cite{DBLP:conf/ijcai/VermaLKBL19}. The pairwise ranking loss is designated for PU-AUC risk minimization. Inspired by~\cite{DBLP:conf/aaai/XieL18}, the risk function in AUC optimization from PU data is equivalent to the risk in supervised AUC optimization. Particularly, let $\ell_{01} (z)$ denote the zero-one loss which returns 1 if $z < 0$, 0.5 if $z = 0$, and 0 otherwise. Supposing that unlabeled data is sampled from a mixture of class distribution $P(\x | y = 0)$ and $P(\x | y = 1)$ completely at random, we show that PU-AUC risk $R_{\mathrm{PU}}$ is an equivalent estimation of PN-AUC risk $R_{\mathrm{PN}}$ as follows.

\begin{equation*}
\begin{aligned}
R_{\mathrm{PU}} =& \underset{\boldsymbol{x} \sim \mathcal{X}_{\mathrm{P}}}{\mathbb{E}}\;\underset{\boldsymbol{x}^{\prime} \sim \mathcal{X}_{\mathrm{U}}}{\mathbb{E}}\ell_{01} \bigl( f(\boldsymbol{x})-f(\boldsymbol{x}^{\prime})\bigr)\\
=& \underset{\boldsymbol{x} \in \mathcal{X}_{\mathrm{P}}}{\mathbb{E}} [\pi  \underset{\overline{\boldsymbol{x}} \sim \mathcal{X}_{\mathrm{P}}}{\mathbb{E}}\ell_{01} \bigl( f(\boldsymbol{x})-f(\overline{\boldsymbol{x}})\bigr) \\& + (1 - \pi)  \underset{\hat{\boldsymbol{x}} \sim \mathcal{X}_{\mathrm{N}}}{\mathbb{E}}\ell_{01} \bigl( f(\boldsymbol{x})-f(\hat{\boldsymbol{x}})\bigr) ]\\
 =& \pi \underset{\boldsymbol{x} \sim \mathcal{X}_{\mathrm{P}}}{\mathbb{E}}\;\underset{\overline{\boldsymbol{x}} \sim \mathcal{X}_{\mathrm{P}}}{\mathbb{E}}\ell_{01} \bigl( f(\boldsymbol{x})-f(\overline{\boldsymbol{x}})\bigr)\\ &+ (1 - \pi) \underset{\boldsymbol{x} \sim \mathcal{X}_{\mathrm{P}}}{\mathbb{E}}\;\underset{\hat{\boldsymbol{x}} \sim \mathcal{X}_{\mathrm{N}}}{\mathbb{E}}\ell_{01} \bigl( f(\boldsymbol{x})-f(\hat{\boldsymbol{x}})\bigr)
\end{aligned}
\end{equation*}
The above equation holds thanks to the linearity of expectation. Note that the first term at the right hand side of the equation is a constant which equals to $\frac{\pi}{2}$. Therefore, it can be omitted in the optimization. Surprisingly, the second term coincides with PN-AUC risk $R_{\mathrm{PN}}$. In other words, we get:
\begin{equation*}
R_{\mathrm{PU}}= (1 - \pi) R_{\mathrm{PN}} +\frac{\pi}{2}
\end{equation*}
It is effortless to see that $R_{\mathrm{PU}}$ is a linear transformation of $R_{\mathrm{PN}}$.
During training, the zero-one loss is usually replaced with a surrogate loss for the convenience of optimization. In our implementation, we substitute it with margin loss defined in Equation~(\ref{equ:pairwiseloss}) which is enough for AUC risk optimization when it gets minimized.

\section{Experiments}
To validate the superiority of~\algo, we conduct experiments on the MNIST\footnote{http://yann.lecun.com/exdb/mnist/}, CIFAR-10\footnote{https://www.cs.toronto.edu/~kriz/cifar.html}, and UCI datasets~\footnote{https://archive.ics.uci.edu/ml/datasets.php}. The comprehensive statistics of used datasets are listed in Table~\ref{exp:dataset}. Notably, the class-ratio of each dataset is the percentage of positive examples among training data.

\begin{table}[ht]
\centering
\begin{tabular}{ L{1.5cm} C{1cm} C{1cm} C{1.2cm} C{1.6cm} }
\toprule
Dataset & \#Train  &  \#Test & \#Feature & Class-ratio\\
 \midrule
ethn & 1,840 & 790 & 30 & 0.50  \\
krvskp & 2,237 & 959 & 36 & 0.49  \\
titanic & 1,540 & 661 & 3 & 0.32 \\
spambase & 3,220 & 1,381 & 57 & 0.40 \\
MNIST & 60,000 & 10,000 & 784 & 0.49 \\
CIFAR-10 & 50,000 & 10,000 & 3,072 & 0.40 \\
\bottomrule
\end{tabular}
\caption{Dataset statistics}\label{exp:dataset}

\end{table}

\subsection{Implementation Details}
Unless otherwise noted, in all experiments we use the multilayer perceptron.  We simply evaluate models using an exponential moving average of their parameters with a decay rate of 0.999. We find in practice that most of~\algo's hyperparameters can be fixed and do not need to be tuned on a per-experiment or per-dataset basis. Specifically, for all experiments, we respectively set the hyperparameters $\beta$ and $\gamma$ the objective function of~\algo~to $1$ and $1$ for simplicity. Further, we only change and $\alpha$ on a per-dataset basis; we found that $\alpha = 1$ are good starting points for tuning.
We used the SGD with nesterov momentum optimizer for all of our experiments. For the experiments in Table 1 and Table 2, we run the experiments for 200 epochs. The initial learning rate was set to $10^{-5}$ on CIFAR-10 and $10^{-3}$ for other datasets. The momentum parameter was set to 0.9. We used a $L_2$ regularization coefficient $10^{-4}$ and a batch-size of 128 in our experiments. All the experiments were done with Pytorch\footnote{https://pytorch.org/}.

\subsection{Competing Methods}
 The following methods are compared:

\begin{itemize}
    \item \emph{Supervised}: This method trains a supervised classifier with lightGBM~\cite{DBLP:conf/nips/KeMFWCMYL17}. It treats unlabeled data as negative and uses hyperopt~\cite{komer2014hyperopt} for hyperparameter optimization.
    \item \emph{WSVM}: The method of~\cite{DBLP:conf/kdd/ElkanN08}. This method treats each unlabeled instance as a combination of positive and negative examples. 
    \item \emph{Ramp}: The method of~\cite{DBLP:conf/nips/PlessisNS14} through optimizing ramp loss. This method is used for comparison with~\algo~on MNIST dataset.
    \item \emph{uPU}: The method of~\cite{DBLP:conf/icml/PlessisNS15} using unbiased PU learning risk estimator.
    \item \emph{nnPU}: The method of~\cite{DBLP:conf/nips/KiryoNPS17} using non-negative unbiased PU learning risk estimator. It is an improved version of uPU which usually overfits because the value of uPU loss can become negative.
    \item \emph{PNU}: The method of~\cite{Sakai2018PNU_AUC} which explicitly optimizes AUC. 
\end{itemize}

\subsection{Results on MNIST}
The model for MNIST is a 3-layer multilayer perceptron (MLP) with ReLU activation function. MNIST has 10 classes originally, and we constructed the $\PP$ and $\NN$ classes from them as follows: MNIST was preprocessed in such a way that 0 constitute the positive class, while one of \{1, 2, 3, 4, 5, 6, 7, 8, 9\} constitutes the negative class separately in each experimental setting. Subsequently, we randomly sample part of $\PP$ which is denoted by $\PP^{\prime}$. We set $\PP = \PP \setminus \PP^{\prime}$ and $\UU = \NN \cup \PP^{\prime}$ to form a PU dataset. 
 We compare Ramp~\cite{DBLP:conf/nips/PlessisNS14} which optimizes ramp loss calculated on positive and unlabeled data with the knowledge of true class-prior. The comparison results are reported in Table~\ref{exp:mnist-error}.  It is noted that~\algo~achieves lowest misclassification rate in comparison with Ramp~\cite{DBLP:conf/nips/PlessisNS14} in most settings we studied. Specifically,~\algo~significantly reduces the classification error rate from 5.48 to 0.33 in 0 vs. 1 task and from 22.58 to 3.51 in 0 vs. 6 task. It indicates that the model initialization for~\algo~produces considerably accurate ``soft'' labels facilitating the consistency loss. In summary, the empirical studies demonstrate that~\algo~is insensitive to class-prior and consistently outperforms Ramp which is one of the representative approaches based on unbiased risk minimization.

\begin{table}[tb]

    \centering
   \ra{1.1}
    \rowcolors{1}{white}{gray!15}
    \begin{tabular}{l C{2cm} |C{1cm} C{1cm} C{1cm} }
    \toprule
     \rule{0pt}{2.2ex} Setting &  Method & 1,200 & 2,400 & 3,600 \\
    \midrule
    \rule{0pt}{2.2ex} \cellcolor{white} & Ramp & 3.36 & 4.85 & 5.48 \\ 
    \multirow{-2}{*}{0 vs. 1}\rule{0pt}{2.2ex} \cellcolor{white} &\algo &  \bf 0.24 & \bf 0.33 & \bf 0.33 \\
    \hline
        \rule{0pt}{2.2ex} \cellcolor{white} & Ramp & 5.15 & 6.96& 7.22 \\ 
    \multirow{-2}{*}{0 vs. 2}\rule{0pt}{2.2ex} \cellcolor{white} &\algo &\bf 4.57 & \bf 2.83 & \bf 2.14
\\
    \hline
        \rule{0pt}{2.2ex} \cellcolor{white} & Ramp & 3.49 & 4.72 & 5.02 \\ 
    \multirow{-2}{*}{0 vs. 3}\rule{0pt}{2.2ex} \cellcolor{white} &\algo &\bf 3.02 & \bf 2.41 & \bf 2.46
 \\
    \hline
        \rule{0pt}{2.2ex} \cellcolor{white} & Ramp & 1.68& 2.05 & 2.21 \\ 
    \multirow{-2}{*}{0 vs. 4}\rule{0pt}{2.2ex} \cellcolor{white} &\algo & \bf 0.76 & \bf 0.66 & \bf 0.46
 \\
    \hline
        \rule{0pt}{2.2ex} \cellcolor{white} & Ramp & 5.21 & 7.22 & \bf 7.46 \\ 
    \multirow{-2}{*}{0 vs. 5}\rule{0pt}{2.2ex} \cellcolor{white} &\algo & \bf 1.60 & \bf 3.63 & 9.62
 \\
    \hline
        \rule{0pt}{2.2ex} \cellcolor{white} & Ramp & 11.47 & 19.87 & 22.58 \\ 
   \multirow{-2}{*}{0 vs. 6} \rule{0pt}{2.2ex} \cellcolor{white} &\algo &\bf 8.57 & \bf 5.31 & \bf 3.51
 \\
    \hline
        \rule{0pt}{2.2ex} \cellcolor{white} & Ramp & 1.89 & 2.55 & 2.64 \\ 
    \multirow{-2}{*}{0 vs. 7}\rule{0pt}{2.2ex} \cellcolor{white} &\algo &  \bf 1.64 & \bf 1.20 & \bf 1.15
 \\
    \hline
        \rule{0pt}{2.2ex} \cellcolor{white} & Ramp & 3.98 & 4.81 & 4.75 \\ 
    \multirow{-2}{*}{0 vs. 8}\rule{0pt}{2.2ex} \cellcolor{white} &\algo &\bf 3.58 & \bf 2.81 & \bf 2.41
 \\
    \hline
        \rule{0pt}{2.2ex} \cellcolor{white} & Ramp & 1.22& 1.60& \bf 1.73 \\ 
    \multirow{-2}{*}{0 vs. 9}\rule{0pt}{2.2ex} \cellcolor{white} &\algo & \bf 1.21 & \bf 0.96 &  3.02
\\
    \bottomrule
    \end{tabular}
        \caption{Misclassification rate (in percent) of~\algo~and Ramp on MNIST dataset. We set the amount of positive data $|\PP|$ from $\{1200, 2400, 3600\}$. The best results are in bold.}\label{exp:mnist-error}
\end{table}

\subsection{Results on CIFAR-10}
We compare our method with state-of-the-art PU learning algorithms on CIFAR-10 dataset. We use the same architecture for all methods as specified in~\cite{DBLP:conf/nips/KiryoNPS17}. CIFAR-10 has 10 classes originally, and we construct the \emph{positive} class and \emph{negative} class as follows. The \emph{positive} class is formed by ``airplane'', ``automobile'', ``ship'', and ``truck'', and the \emph{negative} class is formed by ``bird'', ``cat'', ``deer'', ``dog'', ``frog'', and ``horse''. The results are reported in Table~\ref{cifar10-comparison}. We find that \emph{uPU} is very prone to overfitting and we therefore use a small number of epoch (less than 10). When only 100 positive examples are available, both \emph{nnPU} and \emph{uPU} tend to treat all unlabeled data as negative, which prevent the model from training. Our consistency-regularized model achieve the lowest (best) classification error in 5 out of 6 settings. The comparison result is especially encouraging, considering that \emph{nnPU} and \emph{uPU} use the knowledge of class-prior.

\begin{table}[htbp]
\small
\ra{1.3}
\centering
\begin{tabular}{l|c|c|c|c|c|c}
Method & 100 & 500 & 1k & 2k & 4k & 10k \\
\hline
nnPU & 40.00 & 15.55 &  13.32  & 11.32 & 9.90 &  \bf8.87\\
\hline
uPU & 40.00 & 26.01 &  19.95 & 14.38 & 12.08 & 9.90 \\
\hline
\algo & \bf22.63 & \bf14.00 &  \bf12.21 & \bf11.10 &\bf 9.77 & 8.88  \\
\end{tabular}
\caption{Error rate on \emph{CIFAR-10} dataset with varying number of positively labeled data.}
\label{cifar10-comparison}
\end{table}

\begin{table*}[!t]
    \centering
    \rowcolors{1}{white}{gray!15}
    \begin{tabular}{l C{1.5cm} |C{2cm} C{2cm} C{2cm} C{2cm} C{2cm} }
    \toprule
     \rule{0pt}{2.0ex}Dataset &  Method & 0.01 & 0.05 & 0.1 & 0.2 & 0.4 \\
    \midrule
    \rule{0pt}{2.0ex} \cellcolor{white} & Supervised & 0.50$\pm$0.00 & 0.69$\pm$0.10 & 0.89$\pm$0.02 & 0.92$\pm$0.01 & 0.96$\pm$0.00 \\ 
    \rule{0pt}{2.0ex} \cellcolor{white} &WSVM & 0.56$\pm$0.20 & \bf0.96$\pm$0.01 & \bf0.98$\pm$0.00 & \bf0.99$\pm$0.00 & \bf0.99$\pm$0.00 \\
    \rule{0pt}{2.0ex} \cellcolor{white} & uPU & 0.66$\pm$0.07 & 0.84$\pm$0.04 & 0.94$\pm$0.01 & 0.96$\pm$0.00 & 0.97$\pm$0.00  \\
       \rule{0pt}{2.0ex} \cellcolor{white} & nnPU & 0.63$\pm$0.07 & 0.80$\pm$0.09 & 0.82$\pm$0.03 & 0.86$\pm$0.03 & 0.86$\pm$0.01 \\
          \rule{0pt}{2.0ex} \cellcolor{white} & PNU & 0.71$\pm$0.04 & 0.92$\pm$0.01 & 0.94$\pm$0.01 & 0.95$\pm$0.01 & 0.97$\pm$0.00 \\
         \multirow{-7}{*}{ethn}\rule{0pt}{2.0ex} \cellcolor{white} & \algo & \bf 0.73$\pm$0.04 & 0.92$\pm$0.01 & 0.97$\pm$0.00 & 0.98$\pm$0.00 & 0.94$\pm$0.00 \\
    \hline
    \rule{0pt}{2.0ex} \cellcolor{white} & Supervised & 0.50$\pm$0.00 & 0.81$\pm$0.06 & 0.87$\pm$0.07 & \bf0.97$\pm$0.01 & \bf0.98$\pm$0.00 \\
    \rule{0pt}{2.0ex} \cellcolor{white} &WSVM & 0.61$\pm$0.07 & 0.77$\pm$0.06 & 0.81$\pm$0.05 & 0.85$\pm$0.04 & 0.88$\pm$0.05 \\
    \rule{0pt}{2.0ex} \cellcolor{white} & uPU & 0.72$\pm$0.08 & 0.85$\pm$0.06 & 0.89$\pm$0.04 & 0.95$\pm$0.02 & 0.96$\pm$0.02 \\
       \rule{0pt}{2.0ex} \cellcolor{white} & nnPU & 0.62$\pm$0.09 & 0.78$\pm$0.03 & 0.82$\pm$0.06 & 0.84$\pm$0.02 & 0.88$\pm$0.04 \\
        \rule{0pt}{2.0ex} \cellcolor{white} & PNU & \bf0.72$\pm$0.07 & 0.88$\pm$0.03 & 0.91$\pm$0.03 & 0.95$\pm$0.09 & 0.96$\pm$0.00 \\
         \multirow{-7}{*}{krvskp}\rule{0pt}{2.0ex} \cellcolor{white} & \algo &  0.70$\pm$0.08 & \bf0.90$\pm$0.03 & \bf0.93$\pm$0.02 & 0.96$\pm$0.00 & 0.97$\pm$0.00 \\
     \hline
      \rule{0pt}{2.0ex} \cellcolor{white} & Supervised & 0.50$\pm$0.00 & 0.50$\pm$0.00 & 0.69$\pm$0.02 & 0.72$\pm$0.01 & \bf0.71$\pm$0.00 \\
    \rule{0pt}{2.0ex} \cellcolor{white} &WSVM & 0.35$\pm$0.04 & 0.68$\pm$0.05 & \bf0.73$\pm$0.01 & \bf0.73$\pm$0.02 & 0.71$\pm$0.02 \\
    \rule{0pt}{2.0ex} \cellcolor{white} & uPU & 0.64$\pm$0.09 & \bf0.70$\pm$0.01 & 0.71$\pm$0.00 & 0.71$\pm$0.00 & \bf0.71$\pm$0.00 \\
       \rule{0pt}{2.0ex} \cellcolor{white} & nnPU & 0.63$\pm$0.05 & 0.70$\pm$0.03 & 0.70$\pm$0.02 & 0.71$\pm$0.01 & 0.71$\pm$0.02 \\
       \rule{0pt}{2.0ex} \cellcolor{white} & PNU & 0.63$\pm$0.08 & 0.69$\pm$0.00 & 0.70$\pm$0.02 & 0.70$\pm$0.00 & 0.70$\pm$0.00 \\
         \multirow{-7}{*}{titanic}\rule{0pt}{2.0ex} \cellcolor{white} & \algo & \bf 0.67$\pm$0.01 & 0.68$\pm$0.01 & 0.70$\pm$0.01 & 0.71$\pm$0.01 & \bf0.71$\pm$0.00 \\
\hline
      \rule{0pt}{2.0ex} \cellcolor{white} & Supervised & 0.50$\pm$0.00 & 0.89$\pm$0.02 & 0.90$\pm$0.01 & 0.93$\pm$0.02 & 0.95$\pm$0.01\\ 
    \rule{0pt}{2.0ex} \cellcolor{white} &WSVM & 0.36$\pm$0.01 & 0.58$\pm$0.00 & 0.72$\pm$0.00 & 0.79$\pm$0.00 & 0.85$\pm$0.00 \\
    \rule{0pt}{2.0ex} \cellcolor{white} & uPU & 0.87$\pm$0.05 & 0.91$\pm$0.01 & 0.93$\pm$0.00 & 0.93$\pm$0.01 & 0.94$\pm$0.00 \\
       \rule{0pt}{2.0ex} \cellcolor{white} & nnPU &0.77$\pm$0.07 & 0.87$\pm$0.01 & 0.90$\pm$0.00 & 0.91$\pm$0.01 & 0.92$\pm$0.00 \\
       \rule{0pt}{2.0ex} \cellcolor{white} & PNU &0.76$\pm$0.07 & 0.87$\pm$0.01 & 0.91$\pm$0.01 & 0.93$\pm$0.00 & 0.94$\pm$0.00 \\
         \multirow{-7}{*}{spambase}\rule{0pt}{2.0ex} \cellcolor{white} & \algo &\bf 0.89$\pm$0.01 & \bf0.92$\pm$0.01 & \bf0.94$\pm$0.01 & \bf0.94$\pm$0.01 & \bf0.96$\pm$0.00 \\
      \bottomrule
    \end{tabular}
        \caption{Experimental comparisons on benchmark datasets with varying class-frequency. On each dataset, 10 test runs were conducted. The average AUC and standard deviation are presented. The true value of class-prior is used in uPU and nnPU. The best results in each setting are in bold.~\algo~(ours) achieves competitive performance.}\label{exp:uci-all}
\end{table*}

\subsection{Results on UCI Datasets}
To simulate PU learning problems in the wild, we construct PU data with varying class-frequency $c$. More specifically, we run all competing methods by setting class-frequency $c$ to $c^\prime \in \{0.01, 0.05, 0.1, 0.2, 0.4\}$ through randomly downsampling positive examples and appending them into the unlabeled set $\UU$. For each $c'$, we repeat the experiment $10$ times and report the average performance. The comparison results in terms of AUC are shown in Table~\ref{exp:uci-all}, where means and standard deviations of testing performance based on $10$ random samplings are reported. It is effortless to see that the supervised baseline performs dreadfully with a modest set of positive examples. When more and more positive examples are observed, it achieves competing results because sampled unlabeled examples are becoming more likely to be negative. This validates that the supervised baseline is considerably good and should be compared in PU learning literature. We implement WSVM using Gaussian kernel and it fits the data very well in most cases except when the number of positive examples is extremely limited. Since deep neural networks are used in nnPU, it is no surprise that its performance on small datasets (e.g., ethn, krvskp) is usually worse than other algorithms owing to the lack of labeled examples.  It is interesting to observe that the AUC score of comparison methods is approaching 1.0 even when $c = 0.01$ on \emph{spambase}, which indicates that this dataset is relatively easier to deal with. It is noteworthy that~\algo~is able to achieve superior or comparable results with uPU and nnPU even though they use the true value of class-prior $\pi$ especially when $|\PP|$ is small. In summary, our~\algo~can readily adapted across many practical tasks without the knowledge of class-prior and assumptions over the data distribution.

\subsection{Ablation Studies and Discussion}
In the following, we provide an analysis of the effects of different parts of objective function and RN mining methods.


\begin{table}[htbp]
\small
\ra{1.3}
\centering
\begin{tabular}{l|c|c|c|c}
 Method & 5 & 25 & 50 & 100 \\
\hline
Rand & \bf0.67$\pm$0.01 &  \bf0.68$\pm$0.01 & \bf0.70$\pm$0.01 & \bf0.71$\pm$0.01\\
\hline
Dist & 0.48$\pm$0.14 &  0.53$\pm$0.16 & 0.66$\pm$0.01 & 0.47$\pm$0.15 \\
\hline
NTC & 0.56$\pm$0.11 &  0.68$\pm$0.02 &\bf0.70$\pm$0.01 & 0.69$\pm$0.00 \\
\end{tabular}
\caption{A comparison between negative mining methods on \emph{titanic} dataset with different amount of positive data $|\PP| \in \{5, 25, 50, 100\}$.}\label{rn-comparison}
\end{table}

\subsubsection{How Does the Type of RN Mining Affect Results?}
We report numerical results of employing three different reliable negative mining methods by fixing other components of the networks in Table~\ref{rn-comparison}. Euclidean distance is used in \emph{Dist} method. For \emph{NTC} method, we train a random forest classifier. It is effortless to observe that \emph{Dist} has the worst performance. This indicates that it is unsafe to use distance-based classifiers for unknown data distribution. Therefore, we use \emph{Rand} in all experiments for its observed good performance. We also tried to train the networks without RN examples. Since only positive examples are fed, the networks suffer from overfitting after a few epochs.

\subsubsection{How Much Does Unsupervised Mixup Matter By Itself?}
\begin{figure}[ht]
 \centering
  \begin{subfigure}[tb]{0.48\linewidth}\centering
  \includegraphics[width=1\linewidth]{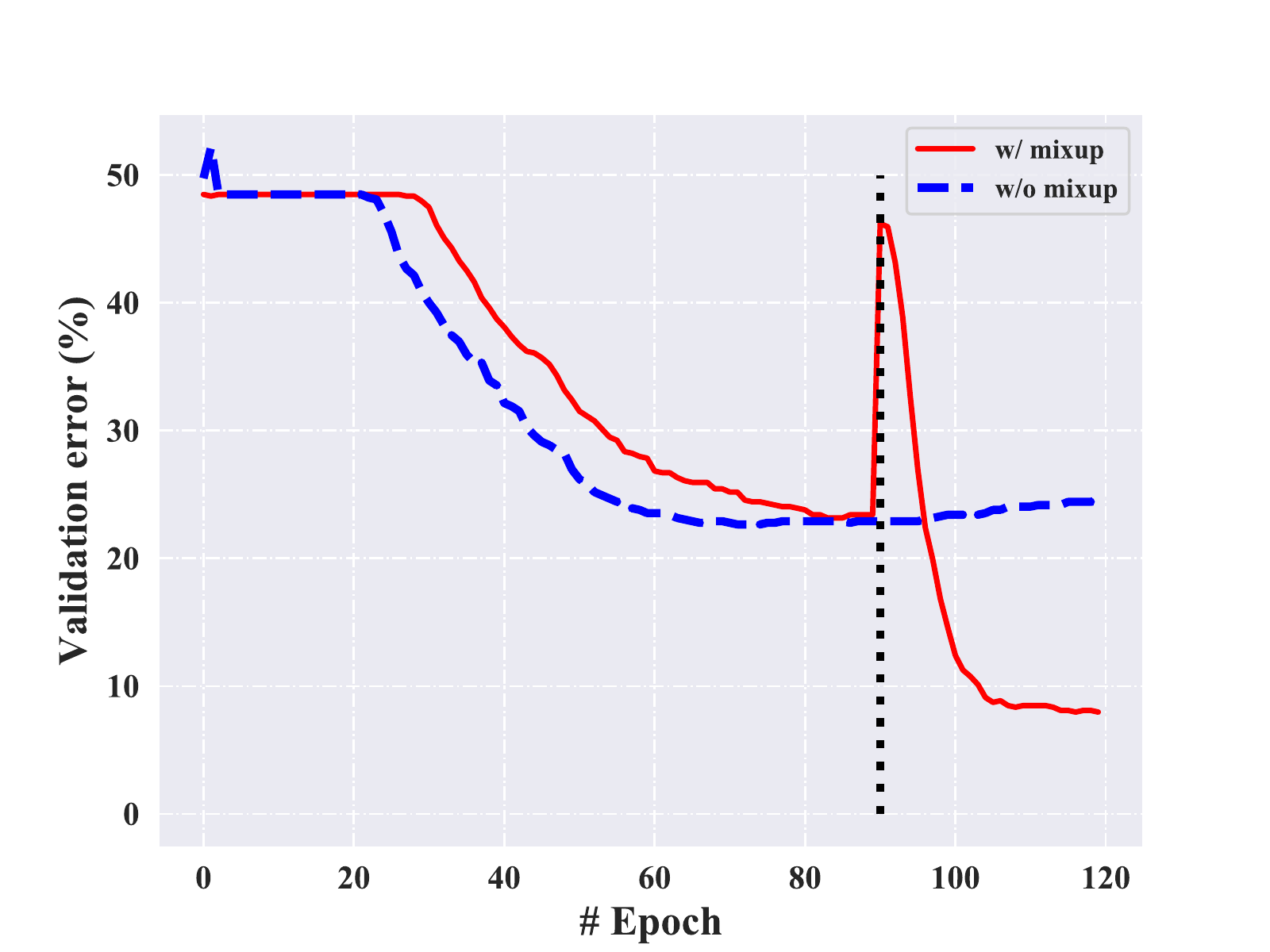}
    \caption{Ablations on mixup}\label{exp:mixup}
  \end{subfigure}
  \begin{subfigure}[tb]{0.48\linewidth}\centering
  \includegraphics[width=1\linewidth]{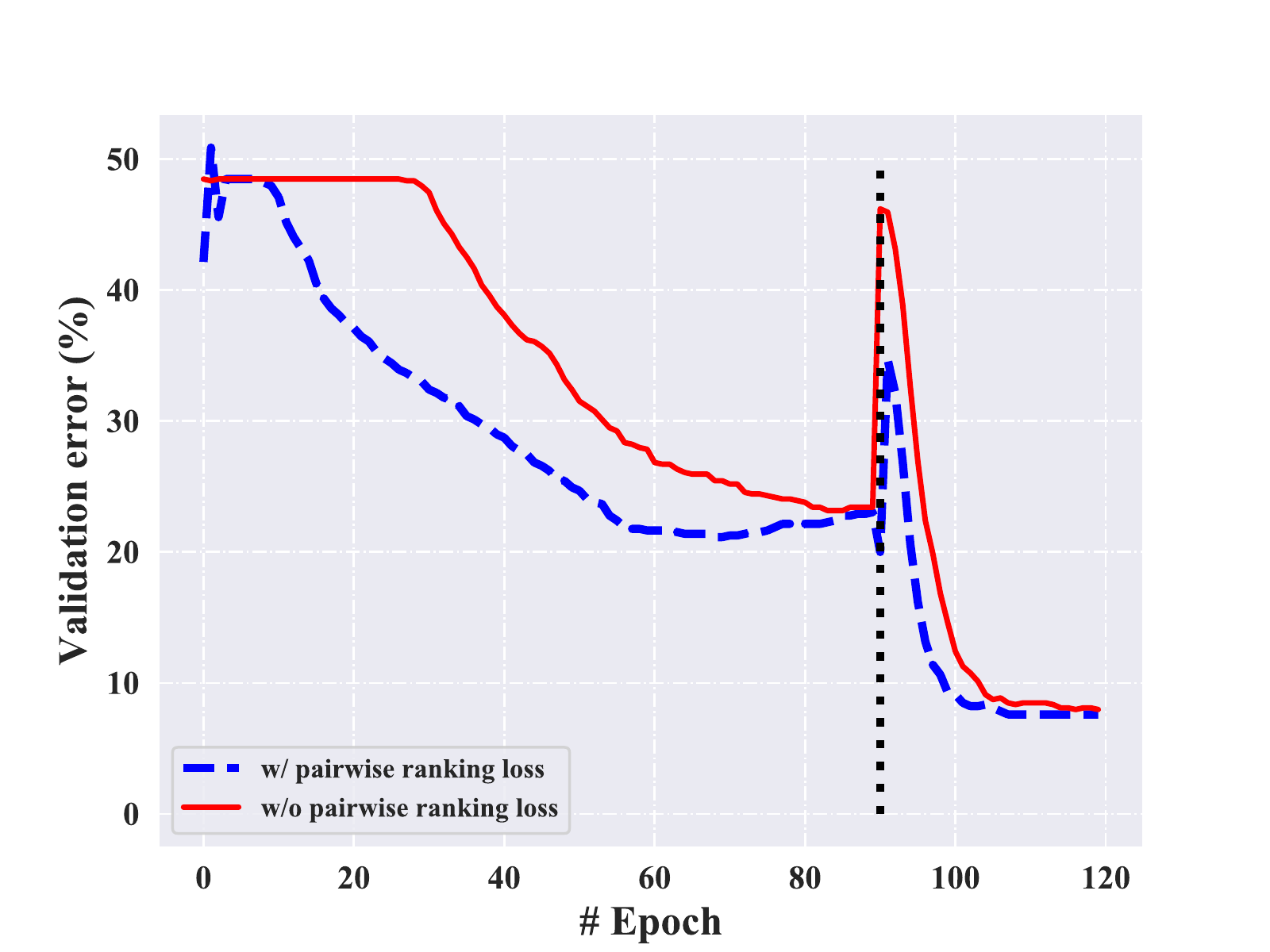}
  \caption{Ablations on margin loss}\label{exp:pairwise}
  \end{subfigure}
  \caption{Ablation studies on unsupervised mixup (left) and margin loss (right).}
  \end{figure}

We study the effect of mixup by training the networks with and without unsupervised interpolation-based consistency loss on one of the image datasets, \emph{ethn}. As shown in Figure~\ref{exp:mixup}, the vertical dashed line indicates the iteration where interpolation training begins. The red and blue lines respectively demonstrate the classification error with and without unsupervised mixup. If we apply the mixup operator on unlabeled data, the misclassification rate initially increases very fast because augmented data is generated which the networks have never seen. After a few iterations, the error rate decreases to less than 10\% which is far smaller than the number without using unsupervised mixup. The results demonstrate the effectiveness of mixup and provide another way of employing unlabeled data for PU learning.

\subsubsection{How Does Margin Loss Affect Results?}
In our objective function, we use margin loss between positive and unlabeled example pairs guiding the model to give higher prediction scores for observed positive examples than unlabeled ones. As shown in Figure~\ref{exp:pairwise}, it results in faster convergence by incorporating margin loss. We also observe that the model has a more stable performance when the mixup is applied as illustrated by the vertical dashed line. The results coincide with our theoretical analysis that the margin loss leads to AUC risk minimization.

In a brief summary, we provide an analysis of the effects of mixup, margin loss, and the negative example mining methods for researchers and practitioners. We find that mixup is more effective on image datasets than other types of data and it also provides a new way of using unlabeled instances for PU learning. Further, by incorporating margin loss, it leads to faster convergence for the networks and makes the model performs more stably when applying mixup. Finally, randomly downsampling unlabeled data as negative is an effective RN mining method for PU learning.

\section{Conclusion}
In this work, we introduce~\algo~which applies interpolation-based consistency regularization to PU learning.~\algo~has two advantages over previous PU learning approaches. First, it does not require the knowledge of class-prior, which otherwise hinders the applicability of the algorithm. Second, through extensive experiments, we find that~\algo~exhibited significantly performance improvements over prior state of the art. Besides, we empirically observe that the proposed negative mining techniques are considerably effective, without which the model corrupts and is unable to incorporate consistency regularization. Specifically, different negative example mining techniques are further investigated and we find the randomized method work very well. We also conduct thorough ablation studies on the consistency regularization and the margin loss. We hope that the proposed consistency regularization will become a standard element in PU learning, and that it will make things easier and simpler for researchers and practitioners.

\bibliographystyle{unsrt}  
\bibliography{references}

\end{document}